\documentclass{article}

\usepackage{microtype}
\usepackage{graphicx}
\usepackage{subcaption}
\usepackage{booktabs} 

\usepackage{hyperref}

\usepackage[accepted]{icml2021}

\usepackage{amsmath}
\usepackage{amssymb}
\usepackage{mathtools}
\usepackage{amsthm}

\usepackage{enumitem}
\usepackage{bbm}
\usepackage{enumitem}
\usepackage{multirow}

\usepackage[capitalize,noabbrev]{cleveref}

\theoremstyle{plain}

\theoremstyle{definition}

\theoremstyle{remark}

\usepackage[textsize=tiny]{todonotes}

\icmltitlerunning{Submission for AdvML-Frontiers 2023}

\begin{document}

\twocolumn[
\icmltitle{Identifying Adversarially Attackable and Robust Samples}



\icmlsetsymbol{equal}{*}

\begin{icmlauthorlist}
\icmlauthor{Vyas Raina}{yyy}
\icmlauthor{Mark Gales}{yyy}
\end{icmlauthorlist}

\icmlaffiliation{yyy}{Machine Intelligence Lab, University of Cambridge, United Kingdom}

\icmlcorrespondingauthor{Vyas Raina}{vr313@cam.ac.uk}
\icmlcorrespondingauthor{Mark Gales}{mjfg@eng.cam.ac.uk}

\icmlkeywords{Adversarial Attacks, Sample Attackability}

\vskip 0.3in
]



\printAffiliationsAndNotice{}  

\begin{abstract}
Adversarial attacks insert small, imperceptible perturbations to input samples that cause large, undesired changes to the output of deep learning models. Despite extensive research on generating adversarial attacks and building defense systems, there has been limited research on understanding adversarial attacks from an input-data perspective. This work introduces the notion of sample attackability, where we aim to identify samples that are most susceptible to adversarial attacks (attackable samples) and conversely also identify the least susceptible samples (robust samples). We propose a deep-learning-based detector to identify the adversarially attackable and robust samples in an unseen dataset for an unseen target model. Experiments on standard image classification datasets enables us to assess the portability of the deep attackability detector across a range of architectures. We find that the deep attackability detector performs better than simple model uncertainty-based measures for identifying the attackable/robust samples. This suggests that uncertainty is an inadequate proxy for measuring sample distance to a decision boundary. In addition to better understanding adversarial attack theory, it is found that the ability to identify the adversarially attackable and robust samples has implications for improving the efficiency of sample-selection tasks. Link to code: \url{https://github.com/rainavyas/img_attackability}
\end{abstract}

\section{Introduction}

Deep learning models have achieved remarkable success across a wide range of tasks and domains, including image classification~\cite{9421942} and natural language processing~\cite{DBLP:journals/corr/abs-1708-05148}. However, these models are also susceptible to adversarial attacks~\cite{https://doi.org/10.48550/arxiv.1412.6572}, where small, imperceptible perturbations to the input can cause large, undesired changes to the output of the model~\cite{DBLP:journals/corr/abs-2008-04094, DBLP:journals/corr/abs-1901-06796, https://doi.org/10.48550/arxiv.2202.10594}. Extensive research has been conducted on generating adversarial attacks and building defense systems~\cite{DBLP:journals/corr/abs-1712-03141, DBLP:journals/corr/abs-1810-00069}, such as detection~\cite{DBLP:journals/corr/abs-1909-06137, Raina_2022, https://doi.org/10.48550/arxiv.1803.10840, DBLP:journals/corr/HendrycksG16b, https://doi.org/10.48550/arxiv.1803.08533} or adversarial training~\cite{https://doi.org/10.48550/arxiv.2203.14046, ijcai2021p591, 9926085}. However, little or no work has sought to understand adversarial attacks from an input-data perspective. In particular, it is unclear why some samples are more susceptible to attacks than others. Some samples may require a much smaller perturbation for a successful attack (\textit{attackable samples}), while others may be less susceptible to imperceptible attacks and require much larger perturbations (\textit{robust samples}). Determining the \textit{attackability} of samples can have important implications for various tasks, such as active learning~\cite{DBLP:journals/corr/abs-2009-00236, 5581075} and adversarial training~\cite{https://doi.org/10.48550/arxiv.2203.14046}. In the field of active learning, adversarial perturbation sizes can be used by the acquisition function to select the most useful/uncertain samples for training~\cite{DBLP:journals/corr/abs-1802-09841, ru-etal-2020-active}. Similarly, adversarial training can be made more efficient by augmenting with adversarial examples for only the most attackable samples to avoid unnecessarily scaling training times, i.e. a variant of weighted adversarial training~\cite{holtz2022learning}.

This work defines the notion of sample attackability: the smallest perturbation size required to change a sample's output prediction. To determine if a particular sample is \textit{attackable} or \textit{robust}, we use the \textit{imperceptibility} threshold in the definition of an adversarial attack. In automated adversarial attack settings, a proxy function is often used to measure human perception~\cite{10.1007/978-3-030-64793-3_10} and thus there can be a range of acceptable thresholds for the imperceptibility boundary, as per the proxy measure. In this work, If a sample's minimum perturbation size is within a \textit{strict} threshold of imperceptibility, then the sample is considered an adversarially attackable sample. In converse, a sample with a perturbation size greater than a much more generous threshold for \textit{imperceptibility} is termed as robust. Further, this work proposes a simple deep-learning based detector to identify the attackable and robust samples, agnostic to a specific model's realisation or architecture. The attackability detector is evaluated on an unseen dataset, for an unseen target model. Further, the attackability detector is evaluated with unmatched adversarial attack methods; i.e. the detector is trained using perturbation sizes defined by the simple Finite Gradient Sign Method (FGSM) attack~\cite{https://doi.org/10.48550/arxiv.1412.6572}, but evaluated on sample perturbation sizes calculated using for example a more powerful Project Gradient Descent (PGD) attack method~\cite{https://doi.org/10.48550/arxiv.1706.06083}. The deep-learning based attackability detector is also compared to alternative uncertainty~\cite{Kim2021EntropyWA, DBLP:journals/corr/abs-2107-03342} based detectors. 

\section{Related Works}

\noindent\textbf{Sample Attackability}:
\citet{DBLP:journals/corr/abs-2010-12989} introduce the notion of sample attackability through the language of \textit{vulnerability} of a sample to an adversarial attack. \citet{Kim2021EntropyWA} further consider sample vulnerability by using model entropy to estimate how sensitive a sample may be to adversarial attacks. This uncertainty approach is considered as a baseline in this work. 

\noindent\textbf{Understanding adversarial samples}: For understanding sample \textit{attackability}, it is useful to consider adversarial attack explanations from the data perspective. Initial explanations~\cite{https://doi.org/10.48550/arxiv.1312.6199, https://doi.org/10.48550/arxiv.1412.5068} argued that adversarial examples lie in large and continuous \textit{pockets} of the data manifold (low-probability space), which can be easily accessed. Conversely, it is also hypothesized~\cite{DBLP:journals/corr/TanayG16} that adversarial examples simply lie in low variance directions of the data, but this explanation is often challenged~\cite{8599715}. Similarly, many pieces of work~\cite{DBLP:journals/corr/abs-1710-10766, DBLP:journals/corr/MengC17, DBLP:journals/corr/LeeHL17, DBLP:journals/corr/abs-1806-00081} argue that adversarial examples lie orthogonal to the data manifold and can so can easily be reached with small perturbations. It can also be argued~\cite{DBLP:journals/corr/abs-1801-02774} that adversarial examples are a simple consequence of intricate and high-dimensional data manifolds.\newline

\noindent\textbf{Use of adversarial perturbations}: In the field of active learning~\cite{DBLP:journals/corr/abs-2009-00236, 5581075}, methods have been proposed to exploit adversarial attacks to determine the minimum perturbation size. The perturbation size is a proxy for distance to a model's decision boundary and thus an acquisition function selects the samples with the smallest perturbations for training the model. However, in these works the perturbation sizes are specific to the model being trained, as opposed to contributing to the overall measure of the \textit{attackability} of a sample, agnostic of the model and adversarial attack method.\newline 

\noindent\textbf{Weighted Adversarial Training}
Knowledge of sample attackability can be applied to the field of adversarial training. Adversarial training~\cite{https://doi.org/10.48550/arxiv.2203.14046} is a popular method to enhance the robustness of systems, where a system is trained on adversarial examples. However, certain adversarial examples can be more \textit{useful} than others and hence certain research attempts have explored reweighing of adversarial examples during training. For example, the adversarial training loss function can be adapted to give greater importance to capture an adversarial example's class margin~\cite{holtz2022learning}. Alternatively, adversarial examples can be re-weighted using the model confidence associated with each example~\cite{DBLP:journals/corr/abs-2010-12989}. However, as opposed to considering the full set of generated adversarial examples, it is more useful to determine which original/non-adversarial samples it is worth using to generate an adversarial example for the purpose of adversarial training. An effective proposed approach~\cite{Kim2021EntropyWA} exploits model uncertainty (such as entropy) associated with each original sample and then adversarial examples need to be generated only for the least certain samples. In the domain of natural language processing, it has been shown that an online meta-learning algorithm can also be used to learn weights for the original samples~\cite{xu-etal-2022-towards}. In contrast to the above methods, this work is the first to use a deep learning approach to predict perturbation sizes for the original/non-adversarial samples, independent of the attack and target model and thus offer a demonstrably more effective method for a variant of weighted adversarial training, labelled active adversarial training (refer to Appendix \ref{sec:active-adv}).  

\section{Adversarial Attacks}

An untargeted adversarial attack is successful in fooling a classification system, $\mathcal F()$, when an input sample $\mathbf{x}$ can be perturbed by a small amount $\boldsymbol{\delta}$ to cause a change in the output class,
\begin{equation}
    \mathcal F(\mathbf x)\neq\mathcal F(\mathbf x + \boldsymbol{\delta}).
\end{equation}
It is necessary for adversarial attacks to be \textit{imperceptible}, such that adversarial perturbations are not easily detectable/noticeable by humans. For images, an imperceptibility constraint is usually enforced using the $l_p$ norm, with $p=\infty$ being the most popular choice, as a proxy to measure human perception,
\begin{equation} \label{eqn:constraint}
    ||\boldsymbol{\delta}||_{\infty} \leq\epsilon,
\end{equation}
where $\epsilon$ is the maximum perturbation size permitted for a change to be deemed appropriately imperceptible. The simplest adversarial attack method for images is Finite Gradient Sign Method (FGSM)~\cite{https://doi.org/10.48550/arxiv.1412.6572}. Here, the perturbation direction, $\boldsymbol{\delta}$, is in the direction of the largest gradient of the output loss function, $\mathcal L()$, for a model with parameters $\boldsymbol{\theta}$,
\begin{equation} \label{eqn:fgsm}
    \boldsymbol{\delta} = \epsilon\texttt{sign}(\nabla_{\mathbf{x}}\mathcal L(\boldsymbol{\theta}, \mathbf x, y)),
\end{equation}
where the $\texttt{sign}()$ functions returns $+1$ or $-1$, element-wise and $y$ is the target class (or the original predicted class by the model). The Basic Iterative Method (BIM)~\cite{DBLP:journals/corr/KurakinGB16a} improves upon the FGSM method, but the most powerful first order image attack method is generally Project Gradient Descent (PGD)~\cite{https://doi.org/10.48550/arxiv.1706.06083}. The PGD method adapts the iterative process of BIM to include random initialization and projection. Specifically, an adversarial example $\mathbf x_0'$ is initialized randomly by selecting a point uniformly on the $l_{\infty}$ ball around $\mathbf x$. Then this adversarial example is updated iteratively,
\begin{equation}\label{eqn:pgd}
    \mathbf x'_{i+1} = \prod_{\mathbf x + \mathcal S} \left[\mathbf x'_{i}+\alpha\texttt{sign}(\nabla_{\mathbf{x}}\mathcal L(\boldsymbol{\theta}, \mathbf x, y))|_{\mathbf{x}=\mathbf x'_i}\right],
\end{equation}
where $\prod$ denotes the projection operator for mapping back into the space of all acceptable perturbations $\mathcal S$ around $\mathbf x$ (element-wise clipping to ensure the elements are within the $l_{\infty}$ ball of size $\epsilon$ around $\mathbf x$) and $\alpha$ is a tunable gradient step. The PGD attack can be run for $t$ iterations. The adversarial perturbation, $\boldsymbol\delta$ is then simply $\boldsymbol\delta=\mathbf x'_{t}-\mathbf x$.

\section{Sample Attackability} \label{sec:sample-robustness}

Sample attackability aims to understand, \textit{how easy is it} to attack a specific sample. For a specific model, $\mathcal F_k$, we can state a sample, $n$'s attackability can be measured directly by the smallest perturbation required to change the classification of the model, 
\begin{equation}\label{eqn:pert}
    \boldsymbol{\hat\delta}^{(k)}_n = \min\limits_{\boldsymbol{\delta}} (\mathcal F_k(\mathbf x_n)\neq\mathcal F_k(\mathbf x_n+\boldsymbol{\delta})).
\end{equation}
A successful adversarial attack requires the perturbation to be imperceptible, as measured by the proxy function in Equation \ref{eqn:constraint}. However, as this is only a proxy measure and there exists variation in what humans deem imperceptible, it is difficult to decide on a single value for $\epsilon$ in the imperceptibility constraint. Hence, in this work, we define sample $n$ as \textit{attackable} for model $k$ if the magnitude of the optimal adversarial perturbation is less than a strict threshold,  $\mathbf{A}_{n,k} = (|\boldsymbol{\hat\delta}^{(k)}_n|<\epsilon_a)$, where any sample that is not attackable can be denoted as $\Bar{\mathbf{A}}_{n,k}$. Conversely, a sample is defined as \textit{robust}, if its adversarial perturbation size is larger than a separate, but more generous (larger) set threshold, $\mathbf{R}_{n,k}=(|\boldsymbol{\hat\delta}^{(k)}_n|>\epsilon_r)$.

However, it is useful to identify samples that are \textit{universally} attackable/robust, i.e. the definition is agnostic to a specific model architecture or model realisation, $k$, used. We can thus extend the definition for \textit{universality} as follows. A sample, $n$, is \textbf{universally attackable} if,
\begin{equation}\label{eqn:uni-att}
    \mathbf A_{n}^{(\mathcal M)} = \bigcap_{k, \mathcal F_k\in\mathcal M} \hspace{0.5em} \mathbf{A}_{n,k},
\end{equation}
where $\mathcal M$ is the set of models in consideration. Similarly a sample is \textbf{universally robust} if, $\mathbf R_{n}^{(\mathcal M)}=\bigcap_{k, \mathcal F_k\in\mathcal M} \hspace{0.5em}\mathbf{R}_{n,k}$. The definition of attackability does not explicitly consider the adversarial attack method used to determine the adversarial perturbations. Experiments in Section \ref{sec:app-corr} demonstrate that the rank correlation of sample perturbation sizes as per different attack methods is extremely high, suggesting that only the thresholds $\epsilon_a$ and $\epsilon_r$ have to be adjusted to ensure the same samples are defined as attackable or robust, independent of the attack method used.

\section{Attackable and Robust Sample Detection} \label{sec:detect}

Section \ref{sec:sample-robustness} defines attackable ($\mathbf{A}_{n,k}$) and robust ($\mathbf{R}_{n,k}$) samples. This section introduces a deep-learning based method to identify the attackable and robust samples in an unseen dataset, for an unseen target model, $\mathcal F_t$. Let the deep-learning attackability detector have access to a seen dataset, $\{\mathbf x_n, y_n\}_{n=1}^N$ and a set of seen models, $\mathcal M = \{\mathcal F_1, \hdots, \mathcal F_{|\mathcal M|}\}$, such that $\mathcal F_t\notin\mathcal M$. Each model can be represented as an encoding stage, followed by a classification stage,
\begin{equation}
    \mathcal{F}_k(\mathbf x_n) = \mathcal F_k^{(\texttt{cl})}(\mathbf h_{n,k}),
\end{equation}
where $\mathbf h_{n,k}$ is the output of the model's encoding of $\mathbf x_n$. A separate attackability detector can be trained for each seen model in $\mathcal M$. For a specific seen model, $k$, we can measure the attackability of each sample using the minimum perturbation size (Equation \ref{eqn:pert}), $\{\boldsymbol{\hat\delta}^{(k)}_n\}_{n=1}^N$. It is useful and efficient to exploit the encoding representation of input images, $\mathbf h_{n,k}$, already learnt by each model. Hence, each deep attackability detector, $\mathcal D_{\theta}^{(k)}$, with parameters $\theta$, can be trained as a binary classification task to determine the probability of a sample being attackable for model $k$, using the encoding at the input,
\begin{equation}
    p(\mathbf{A}_{n,k}) = \mathcal D_{\theta}^{(k)}(\mathbf h_{n,k}).
\end{equation}
 This work uses a simple, single hidden-layer fully connected network architecture for each detector, $\mathcal D$, such that,
\begin{equation}\label{eqn:fcn}
    \mathcal D_{\theta}(\mathbf h) = \sigma(\mathbf{W}_1\sigma(\mathbf W_0\mathbf h)),
\end{equation}
where $\mathbf{W}_0$ and $\mathbf{W_1}$ are the trainable parameters and $\sigma()$ is a standard sigmoid function. Now, we have a collection model-specific detectors that we want to use to determine the probability of a sample being attackable for the unseen target model, $\mathcal F_t$. The best estimate is to use an average over the model-specific detector attackability probabilities,
\begin{equation}
    p(\mathbf{A}_{n,t})\approx \frac{1}{|\mathcal M|}\sum_{k, \mathcal F_k\in\mathcal M}p(\mathbf A_{n,k}).
\end{equation}
However, it is unlikely that this estimate can capture the samples that are attackable specifically for the target model, $\mathcal F_t$'s particular architecture and its specific realisation. Hence, instead it is more interesting to estimate the probability of a \textit{universally attackable} sample (defined in Equation \ref{eqn:uni-att}),
\begin{equation}\label{eqn:uni-det}
    p(\mathbf{A}^{(\mathcal M+t)}_{n})\approx \left[\frac{1}{|\mathcal M|}\sum_{k, \mathcal F_k\in\mathcal M}p(\mathbf A_{n,k})\right ]^{\alpha(\mathcal M)},
\end{equation}
where the parameter $\alpha(\mathcal M)$ models the idea that the probability of sample being universally attackable should decrease with the number of models, i.e. by definition, the number of samples defined as \textit{universally attackable} can only decrease as number of models in $\mathcal M$ increases, as is observed in Figure \ref{fig:sweep} with the \textit{uni} curve lying below all the model-specific curves. Hence, $\alpha(\mathcal M)\geq 1$ and increases with $|\mathcal M|$~\footnote{An alternative product-based model for universal attackability was considered: $p(\mathbf{A}^{(\mathcal M+t)}_{n})\approx \left[\prod_{k, \mathcal F_k\in\mathcal M}p(\mathbf A_{n,k})\right ]^{\alpha(\mathcal M)},$ but empirical results with this method were slightly worse for attackable and robust sample detection.}.

Similarly, detectors can be trained to determine the probability of a sample being universally robust, $p(\mathbf{R}^{(\mathcal M+t)}_{n})$. As a simpler alternative to a deep attackability detector, uncertainty-based detectors can also be used to identify attackable/robust samples. Inspired by \cite{Kim2021EntropyWA}, the simplest form of uncertainty is negative confidence (probability of the predicted class), where it is intuitively expected that the most confident predictions will be for the \textit{robust} samples, $\texttt{conf}_k(\mathbf{x})>\rho_r$ and the least confidence samples can be classed as \textit{attackable}, $\texttt{conf}_k(\mathbf{x})<\rho_a$.

We can evaluate the performance of the attackability detectors on the unseen target model, $\mathcal F_t\notin\mathcal M$, using four variations on defining a sample, $n$ as attackable:
\begin{enumerate}[leftmargin=*]
    \item \textbf{all} - the sample is attackable for the unseen target model.
    \begin{equation}\label{eqn:all}
      \mathbf{A}_{n,t} = (|\boldsymbol{\hat\delta}^{(t)}_n| < \epsilon_a).
    \end{equation}
    \item \textbf{uni} - the sample is universally attackable for the unseen models and the target model.
    \begin{equation}\label{eqn:uni}
        \mathbf{A}^{(\mathcal M+t)}_{n} = \mathbf{A}_{n,t} \cap \mathbf{A}^{(\mathcal M)}_{n}.
    \end{equation}
    \item \textbf{spec} - the sample is attackable for the target model but not universally attackable for the seen models.
    \begin{equation}\label{eqn:spec}
       \mathbf A^{\texttt{spec}}_{n,t} = \mathbf A_{n,t}\cap\bar{\mathbf{A}}^{(\mathcal M)}_{n}.  
    \end{equation}
    \item \textbf{vspec} - a sample is specifically attackable for the unseen target model only.
    \begin{equation}\label{eqn:vspec}
        \mathbf A^{\texttt{vspec}}_{n,t} = \mathbf{A}_{n,t} \cap \left(\bigcap_{k, \mathcal F_k\in\mathcal M} \hspace{0.5em} \bar{\mathbf{A}}_{n,k}\right).
    \end{equation}
\end{enumerate}
As discussed for Equation \ref{eqn:uni-det}, it is expected that the deep learning-based detectors will perform best in the \textbf{uni} evaluation setting.

For an unseen dataset, we can evaluate the performance of attackability detectors using precision and recall. We select a specific threshold, $\beta$, used to class the output of detectors, e.g. $p(\mathbf{A}^{(\mathcal M+t)}_{n})>\beta$ classes sample $n$ as attackable. The precision is $\texttt{prec}=\text{TP}/\text{TP+FP}$ and recall is $\texttt{rec}=\text{TP}/\text{TP+FN}$, where FP, TP and FN are standard counts for False-Positive, True-Positive and False-Negative. A single value summary is given using the F1-score, $\text{F1} = \text{2}*(\texttt{prec}*\texttt{rec})/(\texttt{prec}+\texttt{rec})$. We can generate a full precision-recall curve by sweeping over all thresholds, $\beta$ and then select the best F1 score.

\section{Experiments} \label{sec:exp}
\subsection{Experimental Setup}

Attackability detection experiments are carried out on two standard classification benchmark datasets: Cifar10 and Cifar100~\cite{krizhevsky2009learning}. Cifar10 consists of 50,000 training images and 10,000 test images, uniformly distributed over 10 image classes. Cifar100 is a more challenging dataset, with the same number of train/test images, but distributed over 100 different classes. For both datasets, the training images were randomly separated into a \textit{train} and \textit{validation} set, using a 80-20\% split ratio. For training the attackability detectors, direct access was provided to only the validation data and the test data is used to assess the performance.

Four state of the art different model architectures are considered. Model performances are given in Table \ref{tab:acc}~\footnote{Saved model parameters, hyper-parameter training details and code for all models is provided as a public repository: \url{https://github.com/bearpaw/pytorch-classification}}. Three models (vgg, resnext and densenet) are treated as \textit{seen} models, $\mathcal M$, that the attackability detector has access to during training. The wide-resnet (wrn) model is maintained as an \textit{unseen} model, $\mathcal F_t\notin \mathcal M$ used only to assess the performance of the attackability detector, as described in Section \ref{sec:detect}.

\begin{table}[htb!]
    \centering
    \small
    \begin{tabular}{lrr}
    \toprule
        Model & Cifar10 & Cifar100 \\ \midrule
        vgg-19& 93.3 & 71.8\\
        resnext-29-8-64& 96.2 & 82.5\\
        densenet-121/190-40& 87.5 & 82.7\\ \midrule
        wrn-28-10 & 96.2 &81.6\\
        \bottomrule
    \end{tabular}
    \caption{Model Accuracy (\%)}
    \label{tab:acc}
\end{table}

Two primary adversarial attack types are considered in these experiments: FGSM (Equation \ref{eqn:fgsm}) and the PGD (Equation \ref{eqn:pgd})~\footnote{Other adversarial attacks are considered in Appendix \ref{sec:app-corr}.}. The FGSM attack is treated as a \textit{known} attack type, which the attackability detector has knowledge of during training, whilst the more powerful PGD attack is an \textit{unknown} attack type, reserved for evaluation of the detector. In summary, these experiments use three \textit{seen} models, the validation set and FGSM-based attacks to train an attackability detector. This detector is then required to identify robust/attackable samples for an unseen test set and an unseen target model, perturbation sizes defined as per the known FGSM (\textit{matched} evaluation) or PGD (\textit{unmatched evaluation}).

\subsection{Results}

\begin{figure}[htb!]
    \centering
    \includegraphics[width=\linewidth]{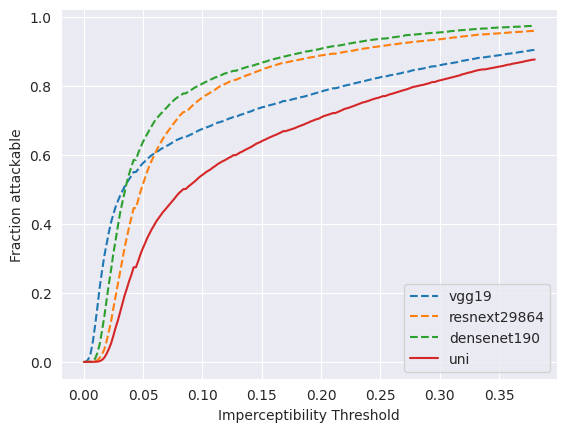}
    \caption{Fraction of \textit{attackable} samples.}
    \label{fig:sweep}
\end{figure}

Initial experiments consider the \textit{matched} evaluation setting. For each \textit{seen} model (vgg, resnext and densenet), the FGSM method is used to determine the minimum perturbation size, $\boldsymbol{\hat\delta}^{(k)}_n$, required to successfully attack each sample, $n$ in the validation dataset for model $k$ (Equation \ref{eqn:pert}). Figure \ref{fig:sweep} shows (for Cifar100 as an example) the fraction, $f$ of samples that are successfully attacked for each model, as the adversarial attack constraint, $\epsilon_a$ is varied:  $f=\frac{1}{N}\sum_n^N\mathbbm{1}_{\mathbf A_{n,k}}$. Based on this distribution, any samples with a perturbation size below $\epsilon_a=0.05$ are termed \textit{attackable} and any samples with a perturbation size above $\epsilon_r=0.39$ are termed \textit{robust}. Note that the \textit{uni} curve in Figure \ref{fig:sweep} shows the fraction of \textit{universally} attackable samples, i.e. these samples are attackable as per all three models (Equation \ref{eqn:uni-att}).


\begin{figure*}[htb!]
     \centering
     \begin{subfigure}[b]{0.24\textwidth}
         \centering
         \includegraphics[width=\textwidth]{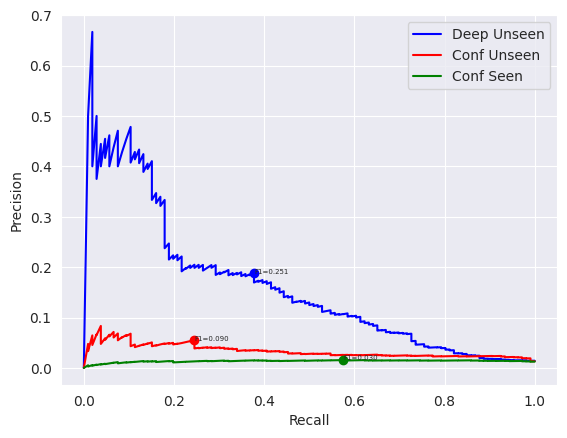}
        \caption{Cifar10-robust}
     \end{subfigure}
     \begin{subfigure}[b]{0.24\textwidth}
         \centering
         \includegraphics[width=\textwidth]{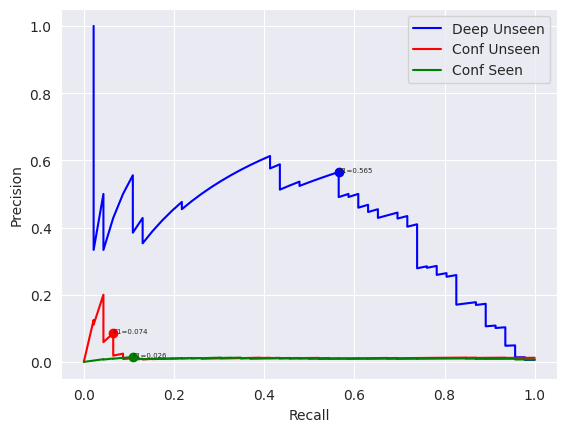}
        \caption{Cifar100-robust}
     \end{subfigure}
 \begin{subfigure}[b]{0.24\textwidth}
         \centering
         \includegraphics[width=\textwidth]{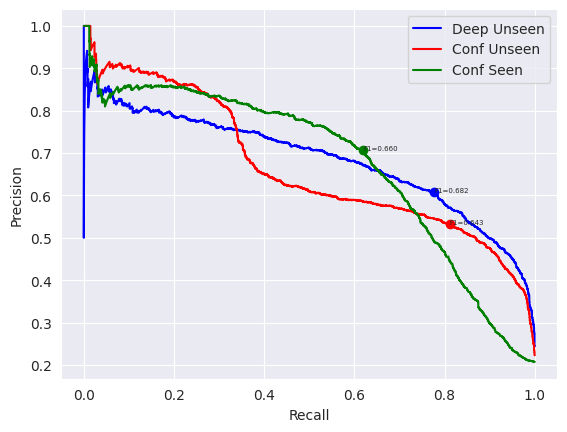}
        \caption{Cifar10-att}
     \end{subfigure}
     \hfill
     \begin{subfigure}[b]{0.24\textwidth}
         \centering
         \includegraphics[width=\textwidth]{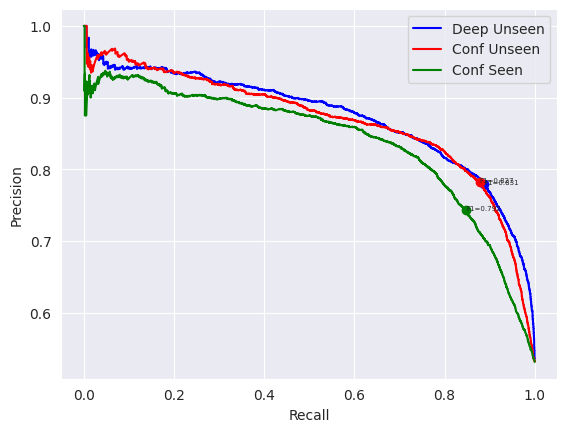}
        \caption{Cifar100-att}
     \end{subfigure}
        \caption{P-R curves for detecting \textit{universal} robust/attackable samples.}
        \label{fig:pr}
\end{figure*}

Section \ref{sec:detect} describes a simple method to train a deep learning classifier to detect robust/attackable samples. Hence, a single layer fully connected network (Equation \ref{eqn:fcn}) is trained with \textit{seen} (vgg, resnext, densenet) models' encodings~\footnote{Encoding stage for each model defined in: \url{https://github.com/rainavyas/img_attackability/blob/main/src/models/model_embedding.py}}, using the validation samples in two binary classification settings: 1) detect attackable samples and 2) detect robust samples. The number of hidden layer nodes for each model's FCN is set to the encoder output size. Training of the FCNs used a batch-size of 64, 200 epochs, a learning rate of 1e-3 (with a factor 10 drop at epochs 100 and 150), momentum of 0.9 and weight decay of 1e-4 with stochastic gradient descent. As described in Section \ref{sec:detect} and inspired by \citet{Kim2021EntropyWA}, an ensemble of the confidence of the three seen architectures, with no access to the target unseen architecture (conf-u) and the confidence of the target architecture, wrn, i.e. having seen the target architecture (conf-s) are also used as uncertainty based detectors for comparison to the deep-learning based (deep) attackability/robust sample detector. To better understand the operation of the detectors, Section \ref{sec:detect} defines four evaluation settings: all (Equation \ref{eqn:all}), uni (Equation \ref{eqn:uni}), spec (Equation \ref{eqn:spec}) and vspec (Equation \ref{eqn:vspec}). Table \ref{tab:m-a} shows the F1 scores for detecting \textit{attackable} samples on the unseen test data for the unseen wrn model, in the matched setting (FGSM attack used to define perturbation sizes for each sample in the test dataset). Note that the scale of F1 scores can vary significantly between evaluation settings as the prevalence of samples defined as \textit{attackable} in a dataset are different for each setting. Table \ref{tab:m-r} presents the equivalent results for detecting robust samples, where the definitions for each evaluation setting update to identifying \textit{robust} samples ($\mathbf R_{n,k}$). For Cifar10 data, the deep unseen detection method performs the best only in the \textit{uni} evaluation setting for both attackable and robust sample detection. This is perhaps expected due to the deep detection method having been designed for this purpose (Equation \ref{eqn:uni-det}), whilst for example the \textit{conf seen} detection method has direct access to the target unseen model (wrn) and is able to perform better in the \textit{spec} and \textit{vspec} settings. When evaluation is on \textit{all} attackable/robust samples for wrn, the superior \textit{vspec} and \textit{spec} performances of the uncertainty detectors, allows them to do better overall. However, for Cifar100 data, the deep detection method is able to perform significantly better in every setting other than the \textit{vspec} evaluation setting, which is expected as this detector has no access to the target model and cannot identify samples specifically attackable/robust for the target model (wrn).

\begin{table}[htb!]
    \centering
    \small
    \begin{tabular}{ll|ccc}
    \toprule
       Setting  & &conf-s & conf-u & deep  \\ \midrule
        \multirow{2}{*}{all} & cifar10 &\textbf{0.693}&0.681&0.683\\
        & cifar100 & 0.845& 0.851&\textbf{0.874}\\ \midrule
        \multirow{2}{*}{uni} & cifar10 &0.660&0.643&\textbf{0.682}\\
        & cifar100& 0.792&0.827 &\textbf{0.831}\\ \midrule
        \multirow{2}{*}{spec} & cifar10 &\textbf{0.239}&0.209&0.206\\
        & cifar100& 0.263&0.261 &\textbf{0.271}\\ \midrule
        \multirow{2}{*}{vspec} & cifar10 &\textbf{0.010}&0.006&0.006\\
        & cifar100&\textbf{0.018} & \textbf{0.018}&\textbf{0.018}\\
        \bottomrule
    \end{tabular}
    \caption{Attackable Sample Detection (F1) in matched setting.}
    \label{tab:m-a}
\end{table}

\begin{table}[htb!]
    \centering
    \small
    \begin{tabular}{ll|ccc}
    \toprule
       Setting  & &conf-s & conf-u & deep  \\ \midrule
        \multirow{2}{*}{all} & cifar10 &0.501 &\textbf{0.502}&0.435\\
        & cifar100 & 0.134& 0.161&\textbf{0.385}\\ \midrule
        \multirow{2}{*}{uni} & cifar10 &0.030&0.090&\textbf{0.251}\\
        & cifar100&0.026 & 0.074&\textbf{0.565}\\ \midrule
        \multirow{2}{*}{spec} & cifar10 &\textbf{0.486}&0.485&0.422\\
        & cifar100& 0.119& 0.141&\textbf{0.286}\\ \midrule
        \multirow{2}{*}{vspec} & cifar10 &\textbf{0.300}&0.288&0.254\\
        & cifar100& 0.042&\textbf{0.061} &0.055\\
        \bottomrule
    \end{tabular}
    \caption{Robust Sample Detection (F1) in matched setting.}
    \label{tab:m-r}
\end{table}

Figure \ref{fig:pr}(a-b) presents the full precision-recall curves (as described in Section \ref{sec:detect}) for detecting robust samples in the \textit{uni} evaluation setting, which the deep-learning based detector has been designed for. It is evident that for a large range of operating points, the deep detection method dominates and is thus truly a useful method for identifying robust samples. Figure \ref{fig:pr}(c-d) presents the equivalent precision-recall curves for detecting attackable samples. Here, although the deep-learning method still dominates over the uncertainty-based detectors, the differences are less significant. Hence, it can be argued that this deep learning-based attackability detector is capable of identifying both attackable and robust samples, but is particularly powerful in detecting robust samples.

In the unmatched evaluation setting the aim is to identify the attackable/robust samples in the test data, where the perturbation sizes for each sample are calculated using the more powerful \textit{unknown} PGD attack method (Equation \ref{eqn:pgd}). PGD attacks used 8 iterations of the attack loop. For each model and dataset, the known FGSM attack and the unknown PGD attacks were used to rank samples in the validation set by the perturbation size, $|\boldsymbol\delta_n|$. In all cases the Spearman Rank correlation is greater than 0.84 for Cifar10 and 0.90 for Cifar100 (Table \ref{tab:corr-pgd}). This implies that the results from the matched setting should transfer easily to the unmatched setting.
\begin{table}[htb!]
    \centering
    \small
    \begin{tabular}{lrrr}
    \toprule
         & vgg & resnext & densenet \\\midrule
        cifar10 & 0.840 &0.849 &0.951\\
        cifar100 & 0.947 & 0.900 & 0.911\\
        \bottomrule
    \end{tabular}
    \caption{Spearman rank correlation (PGD, FGSM) perturbations.}
    \label{tab:corr-pgd}
\end{table}
Table \ref{tab:u} gives the F1 scores for detecting universal attackable/robust samples in the unmatched setting. As the PGD attack is more powerful than the FGSM attack, the definition of the attackable threshold and robustness threshold are adjusted to $\epsilon_a=0.03$ and $\epsilon_r=0.10$. The deep unseen detectors dominate once again (specifically for robust sample detection) and thus, the trends identified for the matched evaluation setting are maintained in the more challenging unmatched setting.

\begin{table}[htb!]
    \centering
    \small
    \begin{tabular}{ll|ccc}
    \toprule
        Uni setting & &conf-s & conf-u & deep  \\ \midrule
        \multirow{2}{*}{Attackable} & cifar10 &0.636&0.754&\textbf{0.777}\\
        & cifar100& 0.846& 0.871&\textbf{0.893}\\ \midrule
        \multirow{2}{*}{Robust} & cifar10 &0.008  & 0.008& \textbf{0.048}\\
        & cifar100& 0.005&0.006 &\textbf{0.233}\\
        \bottomrule
    \end{tabular}
    \caption{Attackable/Robust sample detection (unmatched setting).}
    \label{tab:u}
\end{table}

\section{Conclusion}

This work proposes a novel perspective on adversarial attacks by formalizing the concept of sample attackability and robustness. A sample can be defined as attackable if its minimum perturbation size (for a successful adversarial attack) is less than a set threshold and conversely a sample can be defined as robust if its minimum perturbation size is greater than another set threshold. A deep-learning based attackability detector is trained to identify universally attackable/robust samples for unseen data and an unseen target model. In comparison to uncertainty-based attackability detectors, the deep-learning method performs best, with significant gains for robust sample detection. The understanding of sample attackability and robustness can have important implications for various tasks such as active adversarial training. Future work can explore the significance of attackability on the design of more robust systems.

\section*{Acknowledgements}

This paper reports on research supported by Cambridge University
Press \& Assessment (CUP\&A), a department of The Chancellor, Masters, and Scholars of the University of Cambridge.




\bibliography{main}
\bibliographystyle{icml2021}

\newpage
\appendix
\onecolumn

\section{Unmatched Setting Detailed Results} 

Here more detailed results are presented for detecting attackable (Table \ref{tab:u-a}) and robust samples (Table \ref{tab:u-r}) in an \textit{unmatched} setting, where the adversarial attack method used to measure the sample adversarial perturbation size (PGD) is different from the attack method (FGSM) available during training of the attackability detector. Recall that attackability is defined as the minimum perturbation size in Equation \ref{eqn:pert}.

\begin{table}[htb!]
    \centering
    \begin{tabular}{ll|ccc}
    \toprule
       Setting  & &conf-s & conf-u & deep  \\ \midrule
        \multirow{2}{*}{all} & cifar10 &0.694 &0.776&\textbf{0.797}\\
        & cifar100 & 0.886& 0.882&\textbf{0.907}\\ \midrule
        \multirow{2}{*}{uni} & cifar10 &0.636&0.754&\textbf{0.777}\\
        & cifar100& 0.846& 0.871&\textbf{0.893}\\ \midrule
        \multirow{2}{*}{spec} & cifar10 &0.245&0.245&\textbf{0.261}\\
        & cifar100& \textbf{0.167}& \textbf{0.167}&\textbf{0.167}\\ \midrule
        \multirow{2}{*}{vspec} & cifar10 &\textbf{0.012}&\textbf{0.012} &\textbf{0.012}\\
        & cifar100& \textbf{0.006}& \textbf{0.006}&\textbf{0.006}\\
        \bottomrule
    \end{tabular}
    \caption{Attackable Sample Detection (F1) in unmatched setting.}
    \label{tab:u-a}
\end{table}

\begin{table}[htb!]
    \centering
    \begin{tabular}{ll|ccc}
    \toprule
       Setting  & &conf-s & conf-u & deep  \\ \midrule
        \multirow{2}{*}{all} & cifar10 & 0.195 & \textbf{0.291} & 0.146\\
        & cifar100 &0.088 &0.114 &\textbf{0.307}\\ \midrule
        \multirow{2}{*}{uni} & cifar10 &0.008  & 0.008& \textbf{0.048}\\
        & cifar100& 0.005&0.006 &\textbf{0.233}\\ \midrule
        \multirow{2}{*}{spec} & cifar10 &0.192&\textbf{0.288} &0.144\\
        & cifar100&0.084 & 0.113&\textbf{0.277}\\ \midrule
        \multirow{2}{*}{vspec} & cifar10 &0.149&\textbf{0.221}&0.110\\
        & cifar100&0.041 &0.055 &\textbf{0.095}\\
        \bottomrule
    \end{tabular}
    \caption{Robust Sample Detection (F1) in unmatched setting.}
    \label{tab:u-r}
\end{table}

\section{Extra Unseen Attack Method Experiments} \label{sec:app-corr}

Experiments in the main paper consider the unmatched setting where an attackability detector is trained using FGSM attacked sample perturbation sizes, but evaluated on samples with perturbation sizes defined using the unseen PGD attack method. This section considers further popular adversarial attack methods as unseen attack methods used to measure sample perturbation sizes. Specifically, we consider the Basic Iterative Method (BIM)~\cite{DBLP:journals/corr/KurakinGB16a}, as an attack method with a similar algorithm to the PGD attack method. It is more interesting to understand how sample attackability (minimum perturbation size as per Equation \ref{eqn:pert}) changes when the perturbation sizes are measured using attack methods very different attack algorithms. This may be of interest to a user when trying to understand which samples are susceptible to adversarial attacks in realistic settings, where an adversary may try a range of different attack approaches. Hence, we consider three common whitebox adversarial attack methods to define sample attackability: L-BFGS~\cite{DBLP:journals/corr/TabacofV15}, C\&W~\cite{DBLP:journals/corr/CarliniW16a} and JSMA~\cite{DBLP:journals/corr/PapernotMJFCS15}.

Table \ref{tab:extra-rank} gives the Spearman Rank correlation for minimum perturbation sizes between samples for the FGSM (seen) and the selected (unseen) adversarial attack methods. It is clear that, as was the case with the PGD attack, for Cifar10 and Cifar100, FGSM and BIM attack approaches have a very strong perturbation size correlation. This is perhaps expected as both attack methods are not too dis-similar gradient-based approaches, so will give similar measures of sample attackability. However, the more different attack methods (JSMA, C\&W and L-BFGS) have only a slightly lower rank correlation, suggesting that an attackability detector trained using a simple FGSM attack method can also transfer well in identifying the adversarially attackable and robust samples, where their attackability is defined using very different unseen attack methods.

\begin{table}[htb!]
    \centering
    \begin{tabular}{ll|ccc}
    \toprule
        Attack & Data & vgg & resnext & densenet \\ \midrule
        
        \multirow{2}{*}{BIM} & cifar10 & 0.871 & 0.866 & 0.950\\
        & cifar100 & 0.946 & 0.914 & 0.921\\ \midrule

        \multirow{2}{*}{C\&W} & cifar10 & 0.869 & 0.820 & 0. 945\\
        & cifar100 & 0.928 & 0.919 & 0.910\\ \midrule

        \multirow{2}{*}{L-BFGS} & cifar10 & 0.845 & 0.825 & 0.938\\
        & cifar100 & 0.936 & 0.907 & 0.928\\ \midrule

        \multirow{2}{*}{JSMA} & cifar10 & 0.838 & 0.852 & 0.901\\
        & cifar100 & 0.922 & 0.899 & 0.912\\ 
        \bottomrule
    \end{tabular}
    \caption{Spearman Rank Correlation between FGSM-based and other attack method based sample perturbation sizes (attackability).}
    \label{tab:extra-rank}
\end{table}

\section{Active Adversarial Training}\label{sec:active-adv}

\begin{figure}[htb!]
    \centering
    \includegraphics[width=0.45\textwidth]{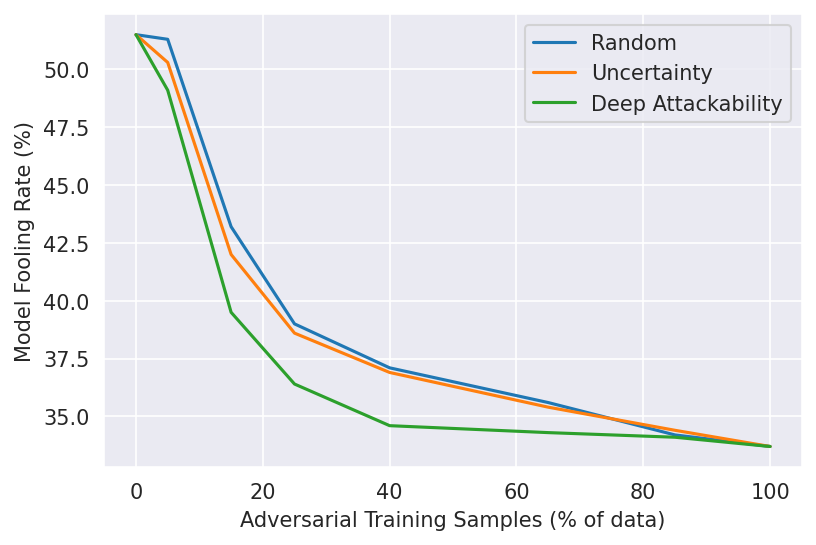}
    \caption{Active Adversarial Training}
    \label{fig:active-adv}
\end{figure}

Adversarial training~\cite{https://doi.org/10.48550/arxiv.2203.14046} is performed by further training trained models on adversarial examples, generated by adversarially attacking original data samples. However, it is computationally expensive to adversarially attack every original data sample. Hence, it is useful to \textit{actively} select a subset of the most \textit{useful} samples for adversarial training. Active adversarial training can be viewed as a strict form of weighted adversarial training~\cite{holtz2022learning}, where adversarial examples are re-weighed in importance during training. Figure \ref{fig:active-adv} shows the robustness (measured by fooling rate)~\footnote{Robustness evaluated on test data attacked using PGD, with $\epsilon=0.03$.} of the target wide-resnet model when adversarially trained using adversarial examples from a subset of the Cifar10 validation data, where the subset is created by different ranking methods: 1) random; 2) a popular \textit{entropy-aware} approach~\cite{Kim2021EntropyWA}, where ranking is as per the uncertainty (entropy) of the trained wide-resent model (uncertainty); and 3) the final ranking method uses the attackability of original validation samples as per the deep attackability detector (unaware of the target wide-resnet model) from this work. The adversarially trained model's robustness is evaluated using the fooling rate on the test Cifar10 data. It is evident that the deep attackability detector gives the most robust model for any fraction of data used for adversarial training. Specifically, with this detector, only 40\% of the samples have to be adversarially attacked for adversarial training to give competitive robustness gains. 


\end{document}